# DeepRacer on Physical Track: Parameters Exploration and Performance Evaluation

Sinan Koparan, Bahman Javadi
School of Computer, Data and Mathematical Sciences
Western Sydney University, Australia
b.javadi@westernsydney.edu.au

## ABSTRACT

This paper focuses on the physical racetrack capabilities of AWS DeepRacer. Two separate experiments were conducted. The first experiment (Experiment I) focused on evaluating the impact of hyperparameters on the physical environment. Hyperparameters such as gradient descent batch size and loss type were changed systematically as well as training time settings. The second experiment (Experiment II) focused on exploring AWS DeepRacer's object avoidance in the physical environment. It was uncovered that in the simulated environment, models with a higher gradient descent batch size had better performance than models with a lower gradient descent batch size. Alternatively, in the physical environment, a gradient descent batch size of 128 appears to be preferable. It was found that models using the loss type of Huber outperformed models that used the loss type of MSE in both the simulated and physical environments. Finally, object avoidance in the simulated environment appeared to be effective; however, when bringing these models to the physical environment, there was a pronounced challenge to avoid objects. Therefore, object avoidance in the physical environment remains an open challenge.

***Keywords – AWS DeepRacer, Reinforcement Learning, Hyperparameters, Autonomous Vehicles, Object Avoidance, Physical Racetrack.***

# Contents

# Background

The transportation field is experiencing a revolution with the rise of autonomous vehicles. Autonomous vehicles have the capability of operating without human intervention. The lack of human intervention is desirable as errors caused by human drivers make up 95 percent of road accidents (Freeman, 2016). These errors occur because of smartphone use, fatigue, or distraction (Kaplan et al., 2015). Pettigrew, Talati, and Norman (2018) highlight the benefits of autonomous vehicles to overcome these errors such as crash prevention and other opportunities such as emission reduction, reduced driving stress, cyclist safety, and mobility for those who are unable to drive. Thus, the integration of autonomous vehicles into modern society can create safe and accessible travel opportunities.

At present, autonomous driving technology can be viewed as being in its early stages. The Society of Automotive Engineers, an international automotive organisation, has defined six different levels of autonomous driving in vehicles (SAE, 2014). These six levels are divided into two groups where the three low levels (level 0 to level 2) require the driver to remain attentive. In the higher levels (level 3 to level 5), drivers will not need to be attentive. Prominent contenders in the autonomous market industry such as Tesla are at level 2 autonomy with their driver assistance system, Autopilot. Tesla drivers are expected to remain attentive while Autopilot is in use (Tesla, 2023). Therefore, autonomous vehicles have not yet achieved fully autonomous capabilities envisioned for the future.

To reach higher levels of autonomous driving, advancements in the domain of reinforcement learning are being made. Kaelbling, Littman, and Moore (1996) describe reinforcement learning as a process where an agent is required to perform actions in an environment based on its state to accumulate the highest rewards. Autonomous vehicles utilise reinforcement learning in some form. For example, reinforcement learning has enabled autonomous vehicles to navigate occluded intersections, control velocity safely, and avoid chain collisions (Isele et al., 2018; Zhu et al., 2019; Muzahid et al., 2022). Therefore, the domain of reinforcement learning must be further researched to attain higher levels of autonomous driving.

Organisations are actively contributing to this endeavour by providing platforms that facilitate learning and research in reinforcement learning. One such organisation is Amazon Web Services (AWS) which offers their DeepRacer service. AWS DeepRacer is a fully autonomous 1/18$^{th}$ scale car that leverages reinforcement learning. It consists of both the AWS DeepRacer console which enables training and evaluating reinforcement learning models in a simulated environment and the AWS DeepRacer vehicle which can run the trained models in a physical environment (AWS, 2023). AWS DeepRacer enables users to practice their reinforcement learning skills in an engaging manner. Thereby, fostering continuous growth and innovation in the field.

# Literature Review

Significant research has been conducted utilising AWS DeepRacer. Revell, Welch, and Hereford (2022) leveraged the AWS DeepRacer service to explore the issue of sim2real performance in the field of robotics where learned skills in a simulated environment does not transfer effectively into the real world. Revell, Welch, and Hereford (2022) found that the best results occurred when a simple action space, simple reward function, and smaller entropy was used. The study by Balaji et al. (2020) also explored sim2real with a focus on domain randomisation. They achieved robust sim2real performance by changing multiple parameters. Additionally, the study by McCalip, Pradhan, and Yang (2023) demonstrated object avoidance in AWS DeepRacer. This was achieved by training three different types of models that utilised different algorithms such as Centreline and two pathfinding algorithms which included A-Star (A*) and Line of Sight (LoS). The A* and LoS outperformed the Centreline model in time per lap and stability. Finally, the study by Ahmed, Ouda, and Abusharkh (2022) utilised the AWS DeepRacer service to explore different hyperparameters in the simulated environment. The results demonstrated that models with a higher learning rate, higher discount factor, and higher entropy were better performing.

While significant research has been conducted utilising AWS DeepRacer, it is essential to acknowledge the limitations of these studies. Many of these studies lacked detailed explanations of their model settings making it difficult for repeatability. The study by Balaji et al. (2020) lacks a discussion on their selected hyperparameters. The study by McCalip, Pradhan, and Yang (2023) focused on object avoidance only in the simulated environment. Similarly, the study by Ahmed, Ouda, and Abusharkh (2022) presented hyperparameter exploration only in the simulated environment. In addition, there was limited exploration of the hyperparameters as there were minimal configurations. Finally, the study by Revell, Welch, and Hereford (2022) explored sim2real focusing mainly on the hyperparameter of entropy. Thus, to add new knowledge, this study will focus on the exploration of other hyperparameters such as gradient descent batch size and loss type in both the simulated and physical environment. Additionally, object avoidance in the physical environment will be explored.

# Experimental Approach

Two separate experiments have been conducted. Experiment I focused on evaluating the impact of hyperparameters on the physical environment. Hyperparameters such as gradient descent batch size and loss type were changed systematically as well as training time settings. Experiment II focused on exploring AWS DeepRacer's object avoidance in the physical environment.

## Experiment I

### Simulation Training and Evaluation

Experiment I involved the training and evaluation of 12 distinct models in both the simulated and physical environment. The configurations of these models are based on the choices for gradient descent batch size, loss type, and training time settings. Appendix A showcases each of these configurations. Prior studies by Ahmed, Ouda, and Abusharkh (2022) and Revell, Welch, and Hereford (2022) have already established optimal values for hyperparameters such as learning rate, discount factor, and entropy. Therefore, these hyperparameters were held constant at their optimal values in the experiment. Other hyperparameters such as the number of epochs and number of experience episodes between each policy-updating iteration remained constant at their default value as they were beyond the scope of this study.

Other settings such as action space, training algorithm, reward function, environment simulation, and race type were also determined for this experiment. Revell, Welch, and Hereford (2022) established that smaller action spaces resulted in better physical racetrack performance. Therefore, an action space containing five turning angles and one speed was selected for this experiment. These values are presented in Figure 1. The training algorithm Proximal Policy Optimisation (PPO) was chosen because Soft-Actor Critic (SAC) is only able to work with continuous action spaces. The reward function that was selected for this experiment is the standard centreline reward function made available by AWS. This reward function allocates higher rewards when the agent is closer to the centreline and minimises rewards when it is further away. The Python code for this reward function is shown in Figure 2. Finally, Re:Invent 2018 has been chosen as the environment simulation and time trial has been chosen as the race type.

| Action | Steering angle | Speed |
|---|---|---|
| 0 | -30.0 degrees | 1.40 m/s |
| 1 | -15.0 degrees | 1.40 m/s |
| 2 | 0.0 degrees | 1.40 m/s |
| 3 | 15.0 degrees | 1.40 m/s |
| 4 | 30.0 degrees | 1.40 m/s |

*Figure 1: Chosen Action Space for Experiment I.*

```python
def reward_function(params):
    '''
    Example of rewarding the agent to follow center line
    '''

    # Read input parameters
    track_width = params['track_width']
    distance_from_center = params['distance_from_center']

    # Calculate 3 markers that are increasingly further away from the center line
    marker_1 = 0.1 * track_width
    marker_2 = 0.25 * track_width
    marker_3 = 0.5 * track_width

    # Give higher reward if the car is closer to center line and vice versa
    if distance_from_center <= marker_1:
        reward = 1
    elif distance_from_center <= marker_2:
        reward = 0.5
    elif distance_from_center <= marker_3:
        reward = 0.1
    else:
        reward = 1e-3  # likely crashed/ close to off track

    return reward
```

*Figure 2: Python Code for Default AWS Centreline Reward Function.*

For each configuration in Appendix A, a model was trained with its appropriate settings. An evaluation in the simulated environment was conducted once training had been completed and the number of off-tracks was recorded. The physical car model was then downloaded and uploaded to the car.

## Physical Environment Evaluation

To evaluate the physical racetrack performance, eight evaluation laps for each configuration were conducted on the physical Re:Invent 2018 racetrack. All models started at Point A (refer to Figure 3). Additionally, the speed setting in the AWS DeepRacer console was set to 28%. In each lap, the number of times that the agent went off track was recorded. If the agent went off track during a lap, it was placed back into the middle of the track from the position that it went off-track.

The top three models with the minimal amount of off-tracks were further evaluated from two other starting positions, Point B and Point C (shown in Figure 3). Eight evaluation laps for the top three models were conducted at both of these starting points.

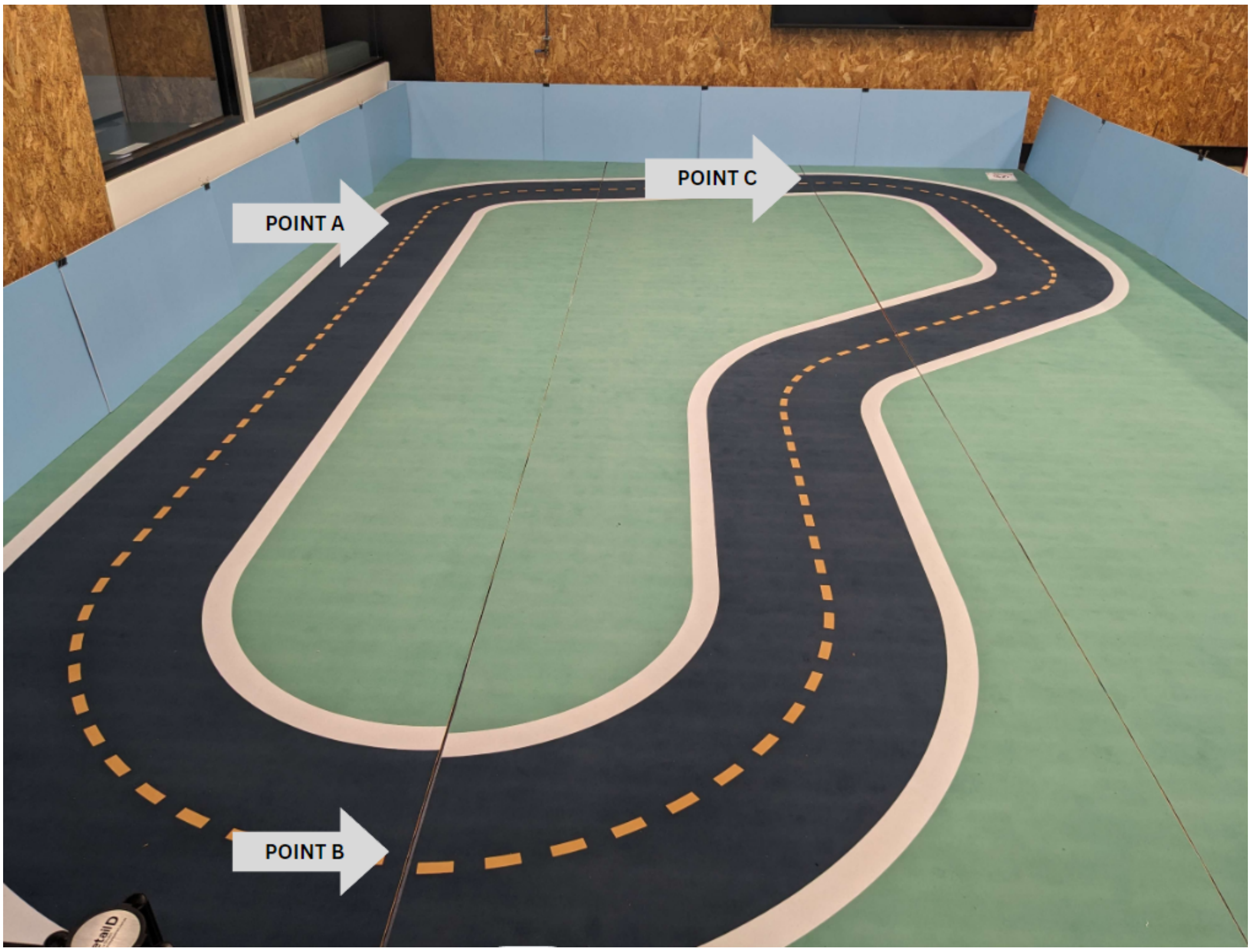

*Figure 3: Point A, Point B, Point C on the Physical Racetrack.*

## Experiment II

### Simulation Training and Evaluation

For Experiment II, three models were trained to determine the capabilities of AWS DeepRacer's object avoidance in the simulated and physical environment. The three models leveraged additional sensors available to AWS DeepRacer such as the LiDAR sensor and stereo cameras. In the simulated environment, these sensors were enabled by building a new car in the AWS DeepRacer garage and toggling on the stereo cameras and LiDAR sensor options. In the physical environment, these sensors were installed into the car.

The first model (Model 1) aimed to test the capabilities of the additional sensors. This model was trained in the simulated environment for one hour using a gradient descent batch size of 512 and a loss type of Huber. Other hyperparameters remained constant at their optimal or default values similar to Experiment I. Other settings such as action space, training algorithm, reward function, environment simulation, and race type remained the same as in Experiment I.

The second model (Model 2) aimed to test AWS DeepRacer's object avoidance. This model was trained in the simulated environment for five hours using a gradient descent batch size of 512 and a loss type of Huber. Settings such as action space, reward function, and race type were changed. The action space for this model was altered to allow the agent to reduce its speed to minimise collisions with objects. The model leveraged the default object avoidance reward function available in AWS. Figure 4 displays the Python code for this reward function. This reward function provides rewards to the agent if it remains within the two borders of the track and allocates a penalty if the agent remains on the same lane as an object while also factoring in the distance of the object. Finally, the race type for this model was the object avoidance race type using the fixed location setting. In the fixed location setting, two objects were chosen and placed at 33% and 85% of the track length. Figure 5 demonstrates the positioning of the two objects. Object 1 was placed on the outside lane and Object 2 was placed on the inside lane.

```python
import math
def reward_function(params):
    '''
    Example of rewarding the agent to stay inside two borders
    and penalizing getting too close to the objects in front
    '''
    all_wheels_on_track = params['all_wheels_on_track']
    distance_from_center = params['distance_from_center']
    track_width = params['track_width']
    objects_location = params['objects_location']
    agent_x = params['x']
    agent_y = params['y']
    _, next_object_index = params['closest_objects']
    objects_left_of_center = params['objects_left_of_center']
    is_left_of_center = params['is_left_of_center']
    # Initialize reward with a small number but not zero
    # because zero means off-track or crashed
    reward = 1e-3
    # Reward if the agent stays inside the two borders of the track
    if all_wheels_on_track and (0.5 * track_width - distance_from_center) >= 0.05:
        reward_lane = 1.0
    else:
        reward_lane = 1e-3
    # Penalize if the agent is too close to the next object
    reward_avoid = 1.0
    # Distance to the next object
    next_object_loc = objects_location[next_object_index]
    distance_closest_object = math.sqrt((agent_x - next_object_loc[0])**2 + (agent_y - next_object_loc[1])**2)
    # Decide if the agent and the next object is on the same lane
    is_same_lane = objects_left_of_center[next_object_index] == is_left_of_center
    if is_same_lane:
        if 0.5 <= distance_closest_object < 0.8:
            reward_avoid *= 0.5
        elif 0.3 <= distance_closest_object < 0.5:
            reward_avoid *= 0.2
        elif distance_closest_object < 0.3:
            reward_avoid = 1e-3  # Likely crashed
    # Calculate reward by putting different weights on
    # the two aspects above
    reward += 1.0 * reward_lane + 4.0 * reward_avoid
    return reward
```

*Figure 4: AWS Object Avoidance Default Reward Function.*

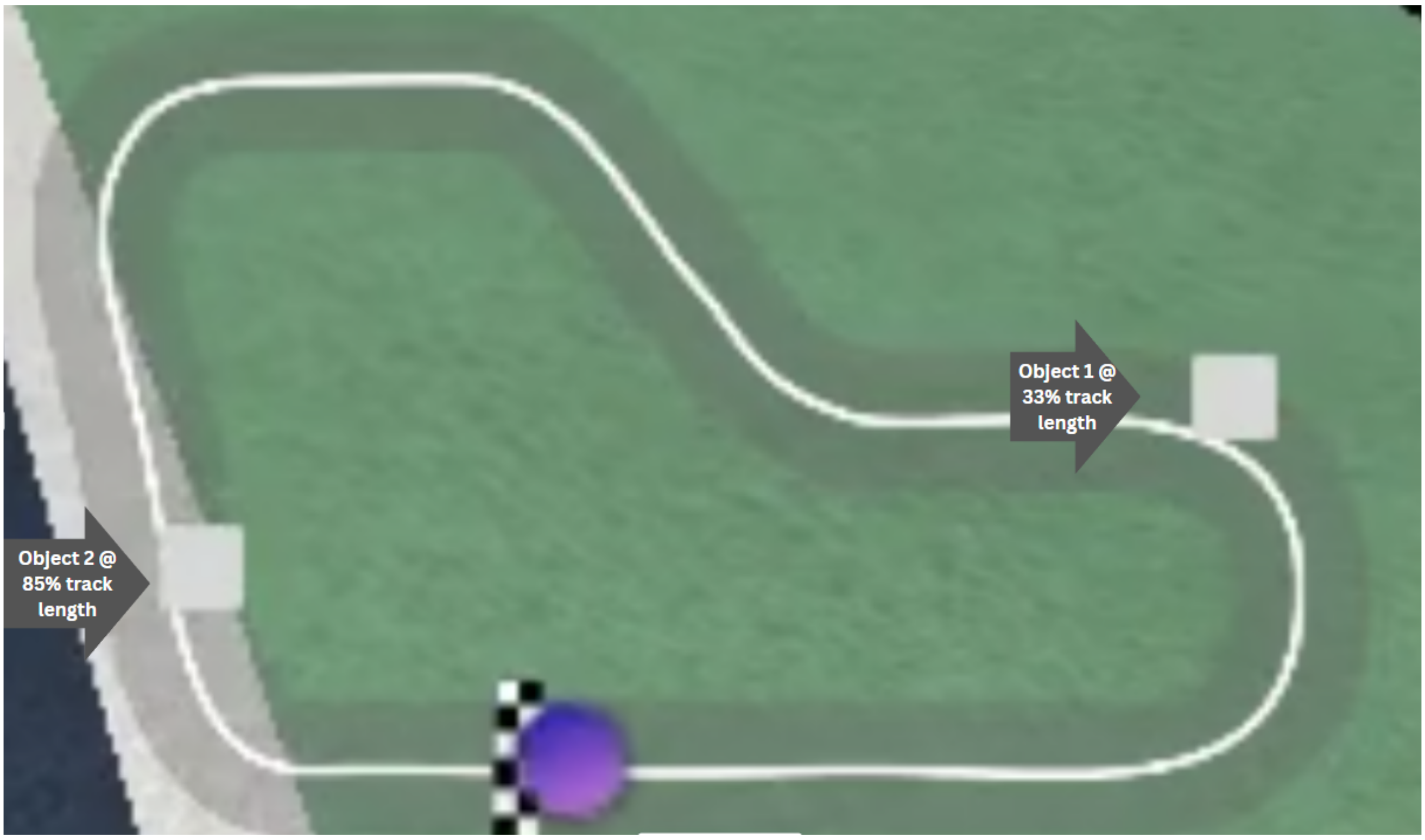

*Figure 5: Positioning of Object 1 and Object 2 on the Racetrack.*

The third model (Model 3) aimed to test AWS DeepRacer's object avoidance with rigorous training. This model was a cloned model of Model 2 that had an additional two hours of training. This model used the same settings as Model 2 with some alterations. The learning rate and gradient descent batch size were lowered to 0.00001 and 64, respectively.

After training the models, an evaluation was conducted. The number of off-tracks, collisions with objects, and time taken were recorded. The physical car model was then downloaded and uploaded to the car.

## Physical Environment Evaluation

To evaluate the physical racetrack performance of the Model 1, three evaluation laps were conducted on the physical Re:Invent 2018 racetrack. The model started at Point A. In each lap, the number of times that the agent went off track was recorded. If the agent went off-track during a lap, it was placed back into the middle of the track from where it went off-track.

To evaluate the capabilities of object avoidance for Model 2 and Model 3 in the physical environment, the agent was positioned 10 yellow stripes before each object. Figure 6 demonstrates the starting position of the agent for its evaluation at Object 1. This procedure was repeated five times for each object. Instances where the agent went off-track were reset and the collisions with objects were recorded.

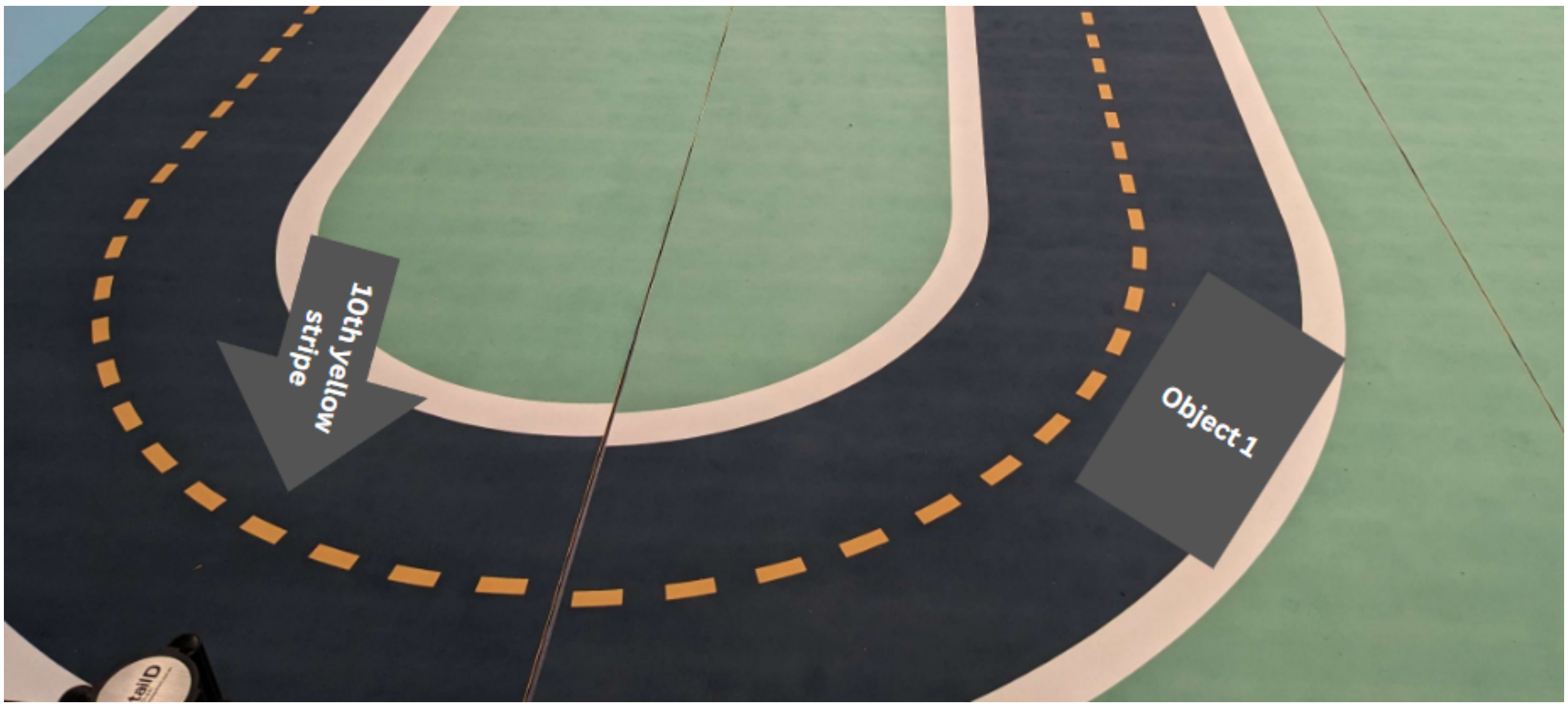

*Figure 6: Agent Starting Position at Object 1.*

# Results and Analysis

## Experiment I – Simulated Environment Results

Table 1 illustrates the results for models trained for an hour in the simulated environment. There is a total of 26 off-tracks. The top performing models include Configuration 3 with 1 off-track and Configuration 6 with 2 off-tracks.

*Table 1: Simulation Results for Models Trained for 1H.*

| Configuration | Gradient Descent Batch Size | Loss Type | Lap 1 Off-Tracks | Lap 2 Off-Tracks | Lap 3 Off-Tracks | Sum |
|---|---|---|---|---|---|---|
| 1 | 64 | Huber | 1 | 2 | 4 | 7 |
| 2 | 128 | Huber | 1 | 1 | 3 | 5 |
| 3 | 512 | Huber | 0 | 0 | 1 | 1 |
| 4 | 64 | MSE | 2 | 2 | 3 | 7 |
| 5 | 128 | MSE | 1 | 2 | 1 | 4 |
| 6 | 512 | MSE | 1 | 0 | 1 | 2 |

Table 2 demonstrates the results for models trained for two hours in the simulated environment. There is a total of 17 off-tracks. The top performing models include Configuration 8, Configuration 9, and Configuration 12 with 1, 1, and 2 off-tracks, respectively.

*Table 2: Simulation Results for Models Trained for 2H.*

| Configuration | Gradient Descent Batch Size | Loss Type | Lap 1 Off-Tracks | Lap 2 Off-Tracks | Lap 3 Off-Tracks | Sum |
|---|---|---|---|---|---|---|
| 7 | 64 | Huber | 2 | 0 | 1 | 3 |
| 8 | 128 | Huber | 0 | 0 | 1 | 1 |
| 9 | 512 | Huber | 1 | 0 | 0 | 1 |
| 10 | 64 | MSE | 2 | 1 | 1 | 4 |
| 11 | 128 | MSE | 2 | 2 | 2 | 6 |
| 12 | 512 | MSE | 1 | 0 | 1 | 2 |

It is evident that extended training time resulted in increased performance in the simulated environment. In addition, models that leveraged the loss type of Huber received fewer off-tracks than those utilising the MSE loss type. Finally, models that used higher values for gradient descent batch size received fewer off-tracks compared to other models that used a lower value for gradient descent batch size.

## Experiment I – Physical Environment Results

Table 3 demonstrates the physical environment results of the models that were trained for an hour. It showcases the number of off-tracks that occurred for each evaluation lap and other statistics such as the sum, mean, standard deviation (SD), and coefficient of variation (CV). Configuration 5 was the only model of the one-hour models that completed an entire lap without any off-tracks. It achieved this on two occasions. In addition, Configuration 5 exhibited the lowest mean among the one-hour models, with a mean value of 2.12. It is also worth noting that Configuration 5 had the highest CV with a value of 64.15%. Finally, the total off-tracks for all the one-hour models are 233.

*Table 3: Physical Environment Results for Models Trained for 1H.*

| Configuration | Lap 1 | Lap 2 | Lap 3 | Lap 4 | Lap 5 | Lap 6 | Lap 7 | Lap 8 | Sum | Mean | SD | CV (%) |
|---|---|---|---|---|---|---|---|---|---|---|---|---|
| 1 | 5 | 3 | 4 | 2 | 2 | 6 | 5 | 5 | 32 | 4 | 1.41 | 35.25 |
| 2 | 3 | 7 | 6 | 3 | 5 | 3 | 3 | 4 | 34 | 4.25 | 1.48 | 34.82 |
| 3 | 8 | 7 | 6 | 3 | 5 | 4 | 6 | 5 | 44 | 5.50 | 1.50 | 27.27 |
| 4 | 7 | 8 | 5 | 7 | 9 | 4 | 6 | 3 | 49 | 6.12 | 1.90 | 31.05 |
| 5 | 4 | 3 | 3 | 2 | 3 | 0 | 0 | 2 | 17 | 2.12 | 1.36 | 64.15 |
| 6 | 12 | 9 | 7 | 7 | 7 | 4 | 5 | 6 | 57 | 7.12 | 2.32 | 32.58 |

Table 4 demonstrates the physical environment results of the models that were trained for two hours. Configuration 8 and Configuration 12 had the least amount of off-tracks of the two-hour models. Their mean values were similar at 3 and 3.25, respectively. The two-hour models had a total of 213 off-tracks.

*Table 4: Physical Environment Results for Models Trained for 2H.*

| Configuration | Lap 1 | Lap 2 | Lap 3 | Lap 4 | Lap 5 | Lap 6 | Lap 7 | Lap 8 | Sum | Mean | SD | CV (%) |
|---|---|---|---|---|---|---|---|---|---|---|---|---|
| 7 | 7 | 4 | 2 | 3 | 5 | 3 | 4 | 4 | 32 | 4 | 1.41 | 35.25 |
| 8 | 1 | 5 | 3 | 3 | 3 | 3 | 4 | 2 | 24 | 3 | 1.12 | 37.33 |
| 9 | 5 | 8 | 7 | 9 | 7 | 5 | 4 | 8 | 53 | 6.62 | 1.65 | 24.92 |
| 10 | 8 | 6 | 2 | 4 | 6 | 6 | 1 | 1 | 34 | 4.25 | 2.49 | 58.59 |
| 11 | 6 | 3 | 3 | 5 | 6 | 8 | 8 | 5 | 44 | 5.50 | 1.80 | 32.73 |
| 12 | 2 | 5 | 3 | 4 | 4 | 2 | 5 | 1 | 26 | 3.25 | 1.39 | 42.77 |

Table 5 demonstrates the physical environment results for the top performing models at Point B. Configuration 5 received the lowest mean with a value of 2.38. Moreover, the CV for Configuration 5 is moderately high at 62.61%. Configurations 8 and 12 received a mean value of 3.25 and 4.62 respectively. The CV for Configurations 8 and 12 were slightly lower than Configuration 5 at 48.00% and 45.89%.

*Table 5: Physical Environment Results for Top Performing models at Point B.*

| Configuration | Lap 1 | Lap 2 | Lap 3 | Lap 4 | Lap 5 | Lap 6 | Lap 7 | Lap 8 | Sum | Mean | SD | CV (%) |
|---|---|---|---|---|---|---|---|---|---|---|---|---|
| 5 | 0 | 2 | 1 | 4 | 3 | 4 | 1 | 4 | 19 | 2.38 | 1.49 | 62.61 |
| 8 | 6 | 2 | 2 | 3 | 1 | 3 | 5 | 4 | 26 | 3.25 | 1.56 | 48.00 |
| 12 | 7 | 6 | 6 | 4 | 0 | 3 | 5 | 6 | 37 | 4.62 | 2.12 | 45.89 |

Table 6 demonstrates the physical environment results for the top performing models at Point C. Configuration 5 received a mean of 2.50. Moreover, the CV for Configuration 5 is high at 74.80%. Configuration 8 received the lowest mean with a value of 1.62. It received a moderate CV of 61.11%. Finally, Configuration 12 received a mean of 2.75 and a high CV of 72.00%.

*Table 6: Physical Environment Results for Top Performing Models at Point C.*

| Configuration | Lap 1 | Lap 2 | Lap 3 | Lap 4 | Lap 5 | Lap 6 | Lap 7 | Lap 8 | Sum | Mean | SD | CV (%) |
|---|---|---|---|---|---|---|---|---|---|---|---|---|
| 5 | 1 | 1 | 2 | 7 | 3 | 3 | 2 | 1 | 20 | 2.50 | 1.87 | 74.80 |
| 8 | 1 | 5 | 3 | 3 | 3 | 3 | 4 | 2 | 13 | 1.62 | 0.99 | 61.11 |
| 12 | 5 | 8 | 7 | 9 | 7 | 5 | 4 | 8 | 22 | 2.75 | 1.98 | 72.00 |

The model that exhibited the minimal amount of off-tracks was Configuration 5. At Point A, it was the only model that was able to complete an entire lap without any off-tracks. Configuration 5 managed to achieve this on two occasions at Point A. It also completed an entire lap at Point B. However, it held the highest CV at Point A, Point B, and Point C, indicating inconsistencies in performance. On the other hand, Configuration 8 demonstrated the most consistency at all points except Point B. It maintained the lowest CV at Point A and Point C. At Point B, Configuration 12 held a lower CV by a difference of 2.11. Both Configuration 5 and Configuration 8 used a gradient descent batch size of 128 and the loss type of Huber with their difference being training time. Thus, it is determined through these results that the optimal gradient descent batch size and loss type are 128 and Huber. However, Configuration 12 demonstrates alternative approaches (gradient descent batch size of 512 and loss type of MSE) for achieving higher performing models.

## Experiment II – Simulated Environment Results

### Model 1 - Time Trial Model with Stereo Cameras and LiDAR Sensor

Table 7 presents the results of the time trial model with the additional sensors. Both laps 1 and 3 had an off-track each, providing a total of 2 off-tracks for the evaluation in the simulated environment. The average time taken to complete a lap is 14.15 seconds.

*Table 7: Simulation Results for the Time Trial Model with Stereo Cameras and LiDAR Sensor.*

| Lap | Time (seconds) | Off Track |
|---|---|---|
| 1 | 14.927 | 1 |
| 2 | 12.592 | 0 |
| 3 | 14.942 | 1 |

### Model 2 - Object Avoidance Model

Table 8 presents the results of the object avoidance model using the additional sensors. This model did not experience any off-tracks. The model only had one collision with an object throughout all the evaluation laps. This collision occurred in lap 2 with Object 1.

*Table 8: Simulation Results for the Object Avoidance Model.*

| Lap | Time (seconds) | Off Track | Crashes |
|---|---|---|---|
| 1 | 15.066 | 0 | 0 |
| 2 | 19.999 | 0 | 1 |
| 3 | 14.299 | 0 | 0 |

### Model 3 - Object Avoidance Cloned Model

Table 9 presents the results of the object avoidance model that was cloned for an additional two hours of training time with the gradient descent batch size and learning rate lowered to 64 and 0.00001, respectively. This model did not go off-track in any of the evaluation laps. Additionally, there was no collisions with objects in any of the evaluation laps.

*Table 9: Simulation Results for the Object Avoidance Cloned Model.*

| Lap | Time (seconds) | Off Track | Crashes |
|---|---|---|---|
| 1 | 13.931 | 0 | 0 |
| 2 | 14.805 | 0 | 0 |
| 3 | 14.598 | 0 | 0 |

The simulation results for the models leveraging the additional sensors were effective. For Model 1, it demonstrated strong capabilities of staying on the track while also leveraging the additional sensors. Model 2 also demonstrated strong capabilities of staying on track. It also showed capabilities of avoiding collisions with objects. However, it did have a collision in lap 2 with Object 1. Finally, Model 3 demonstrated strong capabilities of staying on track and avoiding objects with zero off-tracks and zero collisions. The extended training time for Model 3 appears to have improved its accuracy in avoiding collisions.

## Experiment II – Physical Environment Results

### Model 1 - Time Trial with Stereo Cameras and LiDAR Sensor

Table 10 presents the results of the time trial with the additional sensors in the physical environment. The first lap received eight off-tracks, the second lap received four off-tracks, and the third lap received seven off-tracks.

*Table 10: Physical Environment Results for the Time Trial Model.*

| Lap | Off Track |
|---|---|
| 1 | 8 |
| 2 | 4 |
| 3 | 7 |

### Model 2 - Object Avoidance Model

Table 11 presents the results of the object avoidance model in the physical environment. This model demonstrated minimal capabilities of object avoidance for Object 1. It avoided the object on two occasions. Alternatively, for Object 2, it was unable to detect the object and collided with it on five occasions with a 0% success rate.

*Table 11: Physical Environment Results for Object Avoidance Model.*

| Objects | Collision | Success Rate (%) |
|---|---|---|
| Object 1 (Positioned at 33% track length, outside lane) | 3 | 40 |
| Object 2 (Positioned at 85% track length, inside lane) | 5 | 0 |

### Model 3 - Object Avoidance Cloned Model

Table 12 presents the results of the cloned object avoidance model in the physical environment. This model also demonstrated minimal capabilities for performing object avoidance. For Object 1, the agent avoided the object on one occasion. Similarly, for Object 2, the agent avoided the object on one occasion.

*Table 12: Physical Environment Results for Object Avoidance Cloned Model.*

| Objects | Collision | Success Rate (%) |
|---|---|---|
| Object 1 (Positioned at 33% track length, outside lane) | 4 | 20% |
| Object 2 (Positioned at 85% track length, inside lane) | 4 | 20% |

The physical environment results for the models using the additional sensors were ineffective. Model 1 demonstrated a struggle to stay on track as the number of off-tracks for each evaluation lap was high. In addition, Model 2 and Model 3 highlighted the limited capabilities of performing object avoidance in the physical environment. Both models were only able to avoid objects two occasions.

## Conclusion

In this study, two experiments were conducted utilising AWS DeepRacer. The experiments focused on exploring the effects of hyperparameters and object avoidance in the simulated and physical environments. The first experiment involved systematically changing the gradient descent batch size, loss type and training time settings. It was uncovered that in the simulated environment, models with a higher gradient descent batch size had better performance than models with a lower gradient descent batch size. Alternatively, in the physical environment, a gradient descent batch size of 128 appears to be preferable. It was found that models using the loss type of Huber outperformed models that used the loss type of MSE in both the simulated and physical environment. However, it is evident that alternative hyperparameters exist for achieving higher performing models. Object avoidance in the simulated environment appeared to be effective. However, when bringing these models to the physical environment, there was a pronounced challenge to avoid objects. Therefore, object avoidance in the physical environment remains an open challenge. Future work can focus on determining the appropriate object for the physical racetrack in terms of dimensions, assessing the functionality of the LiDAR sensor, and comparing the results of different sensors. For time trials, future work can focus on exploring different reward functions, optimising speed on the physical racetrack, exploring performance on different racetracks and exploring different action spaces.

# Appendix A

| Configuration | Gradient Descent Batch Size | Number of Epochs | Learning Rate | Discount Factor | Loss Type | Experience Episodes | Reward Function | Entropy | Time (Hours) |
|---|---|---|---|---|---|---|---|---|---|
| 1 | 64 | 10 | 0.01 | 0.999 | Huber | 20 | Centreline | 0.01 | 1 |
| 2 | 128 | 10 | 0.01 | 0.999 | Huber | 20 | Centreline | 0.01 | 1 |
| 3 | 512 | 10 | 0.01 | 0.999 | Huber | 20 | Centreline | 0.01 | 1 |
| 4 | 64 | 10 | 0.01 | 0.999 | MSE | 20 | Centreline | 0.01 | 1 |
| 5 | 128 | 10 | 0.01 | 0.999 | MSE | 20 | Centreline | 0.01 | 1 |
| 6 | 512 | 10 | 0.01 | 0.999 | MSE | 20 | Centreline | 0.01 | 1 |
| 7 | 64 | 10 | 0.01 | 0.999 | Huber | 20 | Centreline | 0.01 | 2 |
| 8 | 128 | 10 | 0.01 | 0.999 | Huber | 20 | Centreline | 0.01 | 2 |
| 9 | 512 | 10 | 0.01 | 0.999 | Huber | 20 | Centreline | 0.01 | 2 |
| 10 | 64 | 10 | 0.01 | 0.999 | MSE | 20 | Centreline | 0.01 | 2 |
| 11 | 128 | 10 | 0.01 | 0.999 | MSE | 20 | Centreline | 0.01 | 2 |
| 12 | 512 | 10 | 0.01 | 0.999 | MSE | 20 | Centreline | 0.01 | 2 |

**Source Code:** https://github.com/SDC-Lab/DeepRacer-Models